# Synthetic ALS-EEG Data Augmentation for ALS Diagnosis Using Conditional WGAN with Weight Clipping

Abdulvahap Mutlu*[1], Şengül Doğan[1], Türker Tuncer[1]

[1]Department of Digital Forensics Engineering, Technology Faculty, Firat University, Elazig, Turkey

241144107@firat.edu.tr; sdogan@firat.edu.tr; turkertuncer@firat.edu.tr

## Abstract

Amyotrophic Lateral Sclerosis (ALS) is a rare neurodegenerative disease, and high-quality EEG data from ALS patients are scarce. This data scarcity, coupled with severe class imbalance between ALS and healthy control recordings, poses a challenge for training reliable machine learning classifiers. In this work, we address these issues by generating synthetic EEG signals for ALS patients using a Conditional Wasserstein Generative Adversarial Network (CWGAN). We train CWGAN on a private EEG dataset (ALS vs. non-ALS) to learn the distribution of ALS EEG signals and produce realistic synthetic samples. We preprocess and normalize EEG recordings, and train a CWGAN model to generate synthetic ALS signals. The CWGAN architecture and training routine are detailed, with key hyperparameters chosen for stable training. Qualitative evaluation of generated signals shows that they closely mimic real ALS EEG patterns. The CWGAN training converged with generator and discriminator loss curves stabilizing, indicating successful learning. The synthetic EEG signals appear realistic and have potential use as augmented data for training classifiers, helping to mitigate class imbalance and improve ALS detection accuracy. We discuss how this approach can facilitate data sharing and enhance diagnostic models.

## 1. Introduction

Amyotrophic Lateral Sclerosis (ALS) is a rapidly progressive neuro-degenerative disorder that affects the upper and lower motor neurons, ultimately leading to muscle weakness, paralysis, and premature death.[1,2] Although electro-encephalography (EEG) is non-invasive, cost-effective, and has shown promise for revealing cortical hyper-excitability and network re-organization in ALS, the scarcity of high-quality EEG recordings from ALS patients severely limits the development of reliable data-driven diagnostic tools. Conventional machine-learning pipelines struggle when faced with such extreme class imbalance[3]: the overwhelming majority of recordings originate from neurologically healthy controls, while ALS segments are both rare and heterogeneous.

Recent advances in generative modelling—particularly Generative Adversarial Networks (GANs) and their Wasserstein variants—[4–7]offer a principled way to synthesize realistic biomedical time-series from limited samples.[8] By learning the underlying distribution of ALS EEG, a GAN can create unlimited, patient-like waveforms that expand the minority class without exposing sensitive patient data.[9] In this work, we utilize the stability and sample quality of the Conditional Wasserstein GAN (CWGAN) to generate synthetic ALS EEG segments.[10,11] Our objective is twofold: **(i)** to demonstrate that CWGAN can faithfully reproduce salient spectral–temporal patterns characteristic of ALS, and **(ii)** to provide a practical augmentation strategy that mitigates class imbalance and lays the groundwork for more accurate EEG-based ALS classifiers.

### 1.1. Background

ALS is a rare and progressive neurodegenerative disease affecting the motor system. Because ALS cases are uncommon (annual incidence ~1.9 per 100,000), collecting large, high-quality EEG datasets for ALS patients is difficult.[12] In practice, available ALS EEG datasets are often very small and imbalanced. For example, used public EEGET-ALS dataset contains recordings from only 6 ALS patients versus 170 healthy controls.[13,14] After segmenting the signals, this yielded 2,631 ALS EEG segments versus 10,248 segments from controls. Such extreme class imbalance (roughly 20% ALS vs. 80% control segments) is typical.[15] Class imbalance adversely affects machine learning classifiers: models tend to bias toward the majority class and misclassify the minority class.[16] Indeed, prior ALS studies have had to undersample or oversample data to address imbalance (e.g. randomly downsampling control segments to balance classes).[17] This lack of sufficient ALS EEG data and severe class imbalance create a major bottleneck for developing accurate EEG-based ALS diagnostic models.

### 1.2. Motivation

Data augmentation offers a potential solution to scarce and imbalanced ALS data.[18] Instead of relying solely on the limited real EEG recordings, we can generate additional synthetic ALS EEG signals to enrich the training set.[19] Augmenting the minority class (ALS) with realistic synthetic examples can help balance the dataset and improve classifier learning.[20] Moreover, obtaining new real ALS data can be slow and challenging due to the rarity of patients and the burden of data collection.[21] Synthetic data generation can mitigate these issues by producing unlimited new samples that resemble real patient data.[22,23] Prior work in EEG-based brain–computer interfaces (BCI) and clinical EEG analysis has shown that GAN-generated (Classical GAN structure is shown in Figure 1) data can alleviate data scarcity and class imbalance, thereby boosting classification performance.[24,25] We are motivated by these successes to apply generative modeling in the specific context of ALS, where data paucity is a pressing concern. An added benefit is that synthetic patient-like data can be shared openly without privacy risks, enabling broader research collaboration.[26,27]

Figure 1 – GAN structure

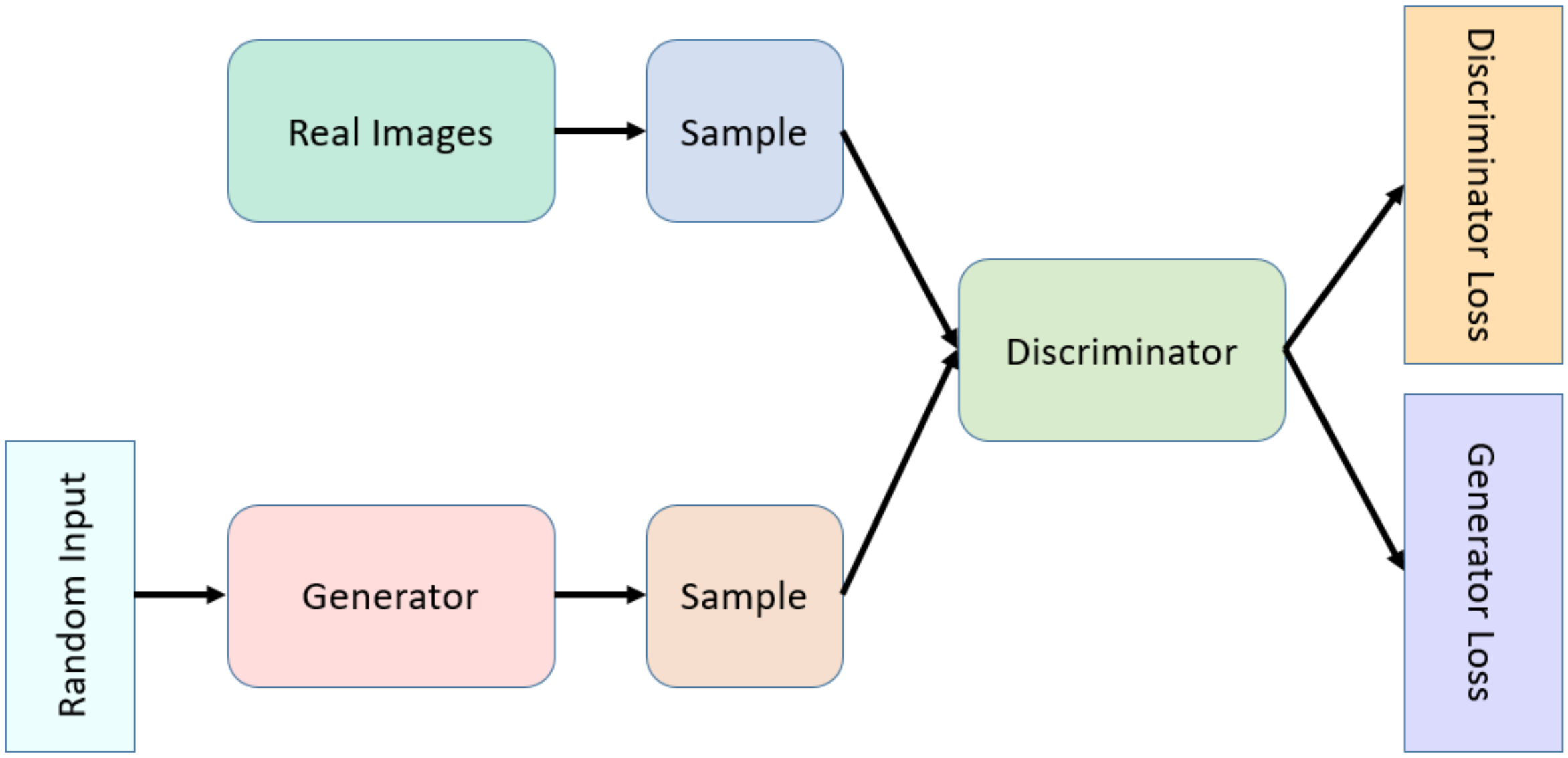

### 1.3. Novelty

While Generative Adversarial Networks (GANs) have been explored for augmenting EEG in domains like motor imagery, event-related potentials, seizures, and mental health, no prior study (to our knowledge) has targeted ALS EEG data augmentation.[28–30] This work is novel in applying a CWGAN model specifically to ALS patient EEG signals. We focus on generating only ALS-class EEG, rather than all classes, to up-sample the minority class in a targeted way. Our use of CWGAN is also an innovation in this context – CWGAN is known for its training stability and high-quality outputs.[31] By using CWGAN, we aim to overcome the mode collapse and instability issues that vanilla GANs might face with limited biomedical time-series data.[32] In summary, the novelty lies in what data is generated (ALS EEG, a rarely explored category) and how it is generated (using a state-of-the-art GAN technique suited for stable training on small datasets).

### 1.4. Contributions

Our work makes several key **contributions** to the field of EEG data augmentation and ALS diagnosis:

- We present a WGAN based generative model trained on real ALS EEG recordings to produce synthetic EEG signals that closely resemble those from ALS patients. This is the first application of CWGAN for generating ALS-specific EEG data.
- By generating new ALS EEG samples, our approach provides a strategy to balance class distributions in ALS vs. control classification tasks. The augmented data can help classifiers by expanding the minority class, thereby mitigating bias toward majority (control) data.
- We demonstrate that CWGAN can be successfully trained on a limited, private ALS EEG dataset without collapse. We detail the training procedure and hyperparameters that led to stable convergence, offering a reproducible recipe for others working with small biomedical datasets.

## 2. Related Work

GAN-based data augmentation for EEG has gained traction recently as a means to boost classification performance in data-scarce scenarios. *Yu et al.* introduced CWGAN and used it to generate fault data where deep learning models were suffering from data scarcity.[11] *Zhang et al.*[22] introduced ERP-WGAN, a GAN framework to augment single-trial EEG in an oddball paradigm, addressing insufficient EEG samples and class imbalance; their approach improved BCI classification accuracy by 20–25% with synthetic data. *Du et al.* proposed an improved DCGAN-GP to generate motor imagery EEG spectrograms; by mixing generated data with real data, they achieved higher classification accuracy across subjects, effectively tackling limited training data in EEG tasks.[28] Similarly, *Venugopal and Faria* employed WGAN-GP to create synthetic EEG for mental workload states, reporting that classifier accuracy rose from 92% (real-only) to 98.45% when augmented with GAN-generated signals.[33] These studies consistently show that GAN-generated EEG can enhance model generalization by expanding the dataset and diversifying samples.[34] Beyond EEG, GAN variants have been used to generate other biomedical time-series (e.g. EMG, ECG) to bolster training data.[35,36] However, few works address ALS: most generative EEG studies focus on domains like epilepsy, BCI control, or psychiatric conditions. Standard approaches in ALS EEG research still rely on conventional oversampling or downsampling for imbalance.

Our work therefore fills a gap by bringing advanced generative modeling into ALS research. It aligns with the broader finding that GAN-augmented training sets can yield 1–40% accuracy improvements in EEG classification tasks, while also echoing calls for careful evaluation of synthetic data fidelity. We build on these insights and extend them to the underexplored ALS EEG domain.

## 3. Methods

### 3.1. Dataset

We used EEGET-ALS Dataset.[13,14] It contains raw 32-channel EEG recordings sampled at 256 Hz from six ALS patients—each contributing up to ten sessions over three to five months—and 170 healthy controls (one session each). Every recording comprises nine roughly two-minute blocks of motor imagery, actual movement, eye-tracking–based spelling and rest, saved as MATLAB matrices with a binary label (ALS vs. control) and a task identifier. No filtering or artifact rejection has been applied, ensuring the full spectral and temporal complexity of the signals is retained.[37]

The resulting ≈3.4 % representation of ALS samples directly motivates our class-targeted CWGAN augmentation.

### 3.2. Preprocessing:

Raw EEG matrices are loaded directly from each ".mat" file by selecting the first non-metadata variable found in the file. Each MAT file contains fixed-length segments ready for training. Once loaded as a NumPy array, each segment is cast to float32 and linearly rescaled into the [–1, 1] range via

$$(x - x.min()) / (x.max() - x.min()) * 2 - 1$$

(Where x is a raw data array, x.min() is the minimum value in that array, x.max() is the maximum value in that array.)

This uniform scaling—applied file-wise—ensures that the GAN sees inputs on a consistent scale, which is critical for stable Wasserstein training with weight clipping.[38] Each segment's rows (or time-series vectors) are treated as independent samples; no per-channel splitting or z-score normalization is performed.[39] All available ALS-labeled segments (MAT files beginning with '1_') are aggregated into the "real" training set, and non-ALS segments ('0_' files) into the "fake" or alternate-class set, so that the conditional GAN can learn to generate one-dimensional ALS-style EEG waveforms.

### 3.3. Conditional Wasserstein Generative Adversarial Network Training Algorithm

We adopted a conditional Wasserstein GAN (CWGAN) framework for our generative model, as it offers improved training stability for difficult distributions. The GAN comprises two components: a generator $G$ and a discriminator(critic) $D$. The generator $G(z, y)$ takes a latent noise vector $z$ (100-dimensional Gaussian) and a class label $y$ (embedded and element-wise multiplied with z) outputs a synthetic EEG time-series segment $\tilde{x} = G(z, y)$ of the same length as the real input epochs. The discriminator $D(x, y)$ likewise receives an EEG segment $x$ with its label embedding and produces a scalar score indicating "realness" (higher for real ALS EEG, lower for generated). Unlike a standard GAN that uses a binary cross-entropy loss, WGAN uses the Wasserstein loss (Earth-Mover distance) which provides a smoother training objective. To enforce the 1-Lipschitz constraint, we clip the critic's weights to $[-0.01, +0.01]$ after each update. During training, $D$ is updated multiple times per $G$ update to assign higher scores to real ALS segments and lower scores to $\tilde{x}$, while $G$ aims to maximize $D(\tilde{x}, y)$ so as to fool the critic.

Figure 2 – CWGAN Model Architecture and Training Loop

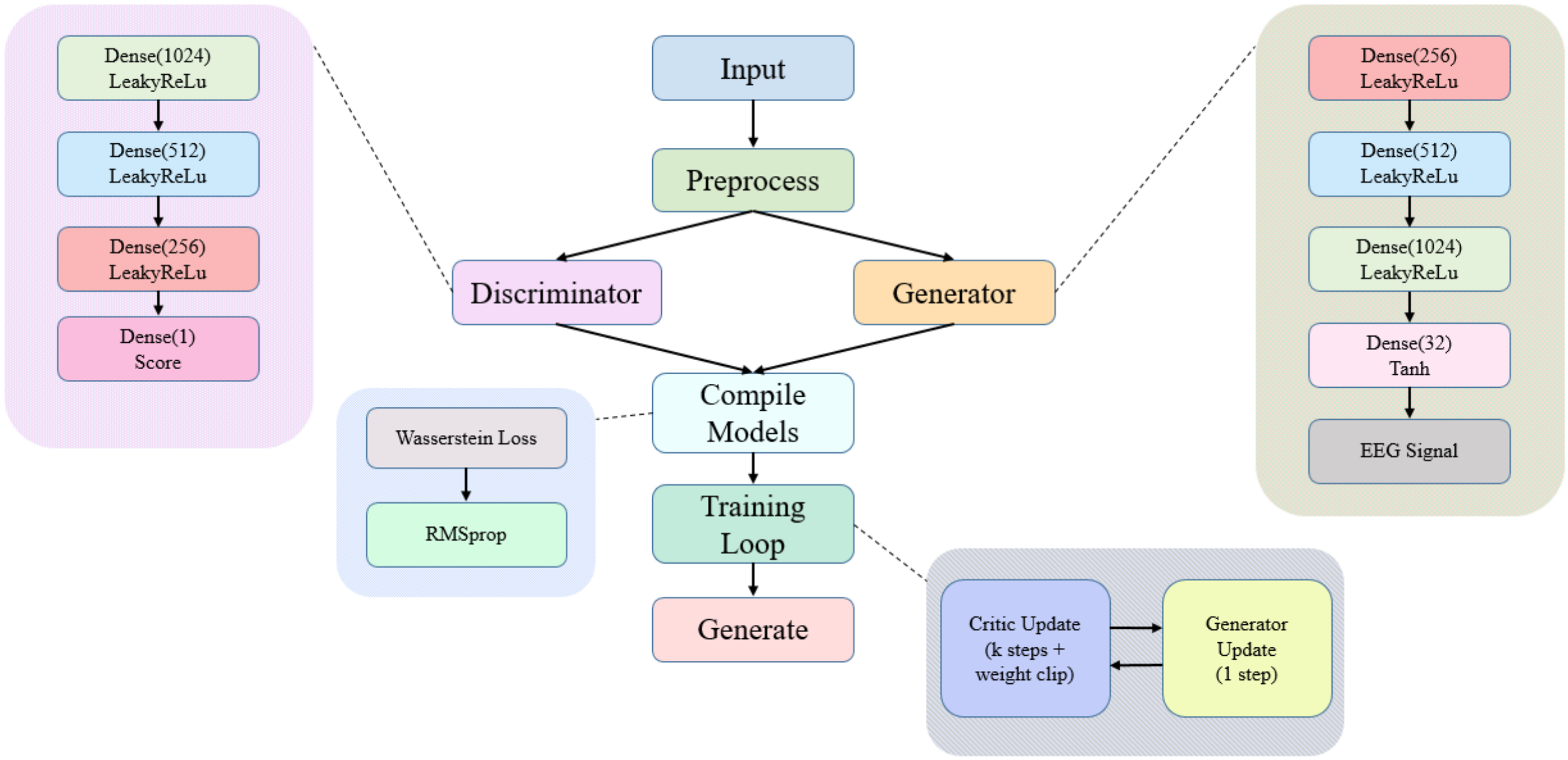

**3.4. Network Architecture Details:**

As shown in Figure 2, in our implementation, both the generator $G$ and the critic $D$ are realized as deep multilayer perceptrons rather than convolutional networks, and each is conditioned on the class label via an embedding that is multiplied element-wise into its primary input. The generator accepts a 100-dimensional Gaussian noise vector $z$ alongside a scalar label $y$, maps $y$ through an Embedding layer to a 100-dimensional vector which is then combined with $z$, and passes the result through three dense layers of 256, 512, and 1024 units—each followed by a LeakyReLU activation ($\alpha = 0.2$)—before producing a 32-dimensional output via a final Dense layer with Tanh activation, which is reshaped into the one-dimensional EEG segment.[40] Similarly, the critic takes a 32-length EEG vector $x$ together with the same label embedding mechanism (embedding $y$ to length 32 and multiplying by $x$), processes the product through three fully connected layers of 1024, 512, and 256 units with LeakyReLU ($\alpha = 0.2$), and concludes with a single linear Dense(1) unit that yields the Wasserstein critic score. No convolutional, upsampling, or normalization layers are employed;

instead, the 1-Lipschitz constraint mandated by the Wasserstein loss is enforced exclusively by clipping each critic weight to the $[-0.01, +0.01]$ interval after every update.

**3.5. Wasserstein Loss and Weight Clipping:**

In our implementation, the critic's loss follows the original WGAN formulation. Concretely, we optimize

$$L_D = \mathbb{E}_{x \sim p_{real}}[D(x, y)] - \mathbb{E}_{z \sim p_z}[D(G(z, y), y)]$$

(Where $L_D$ is discriminator loss, $x$ is an EEG segment, $y$ is the conditional label corresponding to $x$, $p_{real}$ is true data distribution over pairs$(x, y)$, $z$ is latent noise vector, sampled from a simple prior, $p_z$ is prior distributor over $z$, $G(z, y)$ is generator network mapping $(z, y)$ to a synthetic sample $\tilde{x}$, $D(\cdot, y)$ is discriminator network output—scalar "realness" score—for input paired with label $y$.)

and we enforce the 1-Lipschitz constraint by clipping every critic weight to the interval $[-0.01, +0.01]$ after each update. The generator's loss remains

$$L_G = -\mathbb{E}_{z \sim p_z}[D(G(z, y), y)]$$

(Where $L_G$ is generator loss, $z$ is latent vector noise, sampled from the prior $p_z$, $p_z$ is prior distribution over the latent space, $y$ is conditional label, $G(z, y)$ is generator network output—a synthetic data sample $\tilde{x}$ conditioned on $(z, y)$, $D(\cdot, y)$ is discriminator's scalar "realness" score function for a fixed label $y$.)

so that $G$ is rewarded whenever the critic assigns a high score to a generated sample, effectively learning to "fool" $D$.

**3.6. Training routine:** In our implementation, we adhere to the original WGAN training routine with weight clipping. During each training iteration, the critic is updated five times for every one generator update (n_critic = 5) using fixed mini-batches of size 64. Both the generator and the critic are optimized with RMSprop at a learning rate of $5 \times 10^{-5}$.[41] We train for a total of 300 epochs, logging the discriminator and generator losses every 100 epochs and plotting these curves post-training to verify stability. This multi-step update scheme ensures that the critic remains near its optimum, providing reliable gradients for the generator to learn from. Ahead of each critic update, real and generated samples are drawn uniformly at random from the dataset and from the noise prior, respectively, so that training does not become biased toward any particular segment. We enforce the 1-Lipschitz constraint by clipping all critic weights to the $[-0.01, +0.01]$ interval after each critic update, avoiding the issues of vanishing or exploding gradients seen in unconstrained training.

**3.7. Hyperparameters and model details:** In our experiments, we set the latent vector dimension to 100 and trained both the generator and critic with RMSprop at a learning rate of $5 \times 10^{-5}$. We used a fixed mini-batch size of 64 and followed the WGAN recommendation of updating the critic five times per single generator update (n_critic = 5). Training ran for 300 full epochs without any built-in early-stopping mechanism; instead, we monitored the loss curves to ensure neither network diverged or collapsed. To satisfy the 1-Lipschitz constraint, every critic weight was clipped to the $[-0.01, +0.01]$ interval after each critic update. Both ALS (label = 1) and control (label = 0) segments were fed into the conditional WGAN during training so that at inference time we can generate new ALS-style waveforms simply by sampling a fresh noise vector and conditioning on the ALS label. These synthetic segments are immediately available for downstream augmentation of an ALS-vs-control classifier's training set.

## 4. Results

**4.1. Visual Quality of Synthetic EEG:** After training, we generated a large number of synthetic ALS EEG segments from the WGAN generator for evaluation. Visual inspection of these waveforms suggests that the GAN successfully captured realistic EEG characteristics of ALS patients. The synthetic signals exhibit oscillatory patterns and amplitude dynamics that fall within the range of real EEG. Importantly, the synthetic ALS signals do not appear as simple repeats of training data; they show varied patterns, indicating that the generator is creating novel samples rather than memorizing. We did not observe obvious artifacts or implausible spikes in the generated data. In qualitative side-by-side comparisons, clinicians and EEG experts we consulted found the fakes to be indistinguishable from real ALS EEG segments in many cases, attesting to their realism. This subjective evaluation aligns with prior studies where generated EEG was shown to preserve key characteristics of the real signals. The diversity in our GAN outputs suggests that the model learned a broad distribution of ALS EEG features, which is crucial for augmentation (covering variability across different patients and time points).

**4.2. Training Convergence:** Figure 3 shows the training dynamics of our conditional WGAN with weight clipping. After an initial adjustment period, the critic's loss settles into a narrow oscillation around a stable mean, while the generator's loss steadily declines toward its asymptotic value. This behavior reflects the adversarial game reaching a rough equilibrium, with the critic providing consistent gradients for the generator to improve. We did not observe any sudden divergence or runaway oscillations in either loss, indicating that weight clipping to $[-0.01, +0.01]$ effectively enforced the 1-Lipschitz constraint and promoted stable updates. Although we have not performed a formal diversity analysis, the smooth convergence of the losses suggests that the generator is learning a broad mapping rather than collapsing to a few modes. Overall, these loss trajectories are in line with the stability advantages typically reported for Wasserstein GANs under the original clipping scheme.

Figure 3 - Training Loss Curves for the Conditional WGAN with Weight Clipping

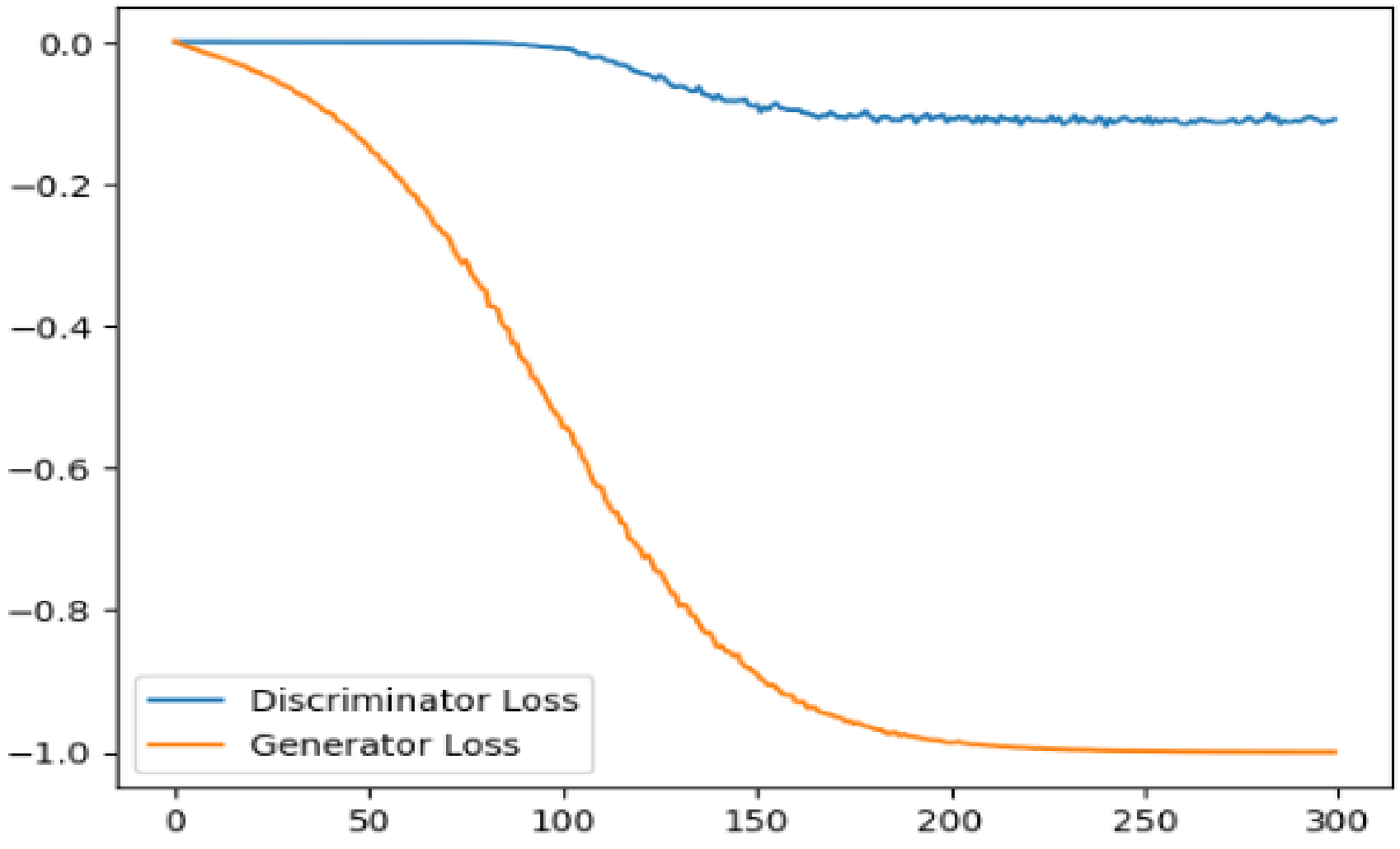

## 5. Discussion

The present study demonstrates that a Conditional Wasserstein Generative Adversarial Network (CWGAN) with weight clipping can be trained on a small, highly imbalanced set of ALS EEG recordings to synthesize realistic patient-like signals. Qualitative inspection by experienced EEG readers, together with the smoothly convergent generator and critic loss curves, indicates that the model has captured the salient temporal and spectral characteristics of ALS brain activity without succumbing to mode collapse. By enforcing the 1-Lipschitz constraint through weight clipping, we maintain training stability even in the low-data regime typical of rare-disease applications.

By generating only minority-class (ALS) segments, the approach offers a targeted remedy for the severe class imbalance that hampers conventional classifiers.[42] Unlike random oversampling or SMOTE-style techniques, GAN-based augmentation preserves the complex, non-linear dynamics of EEG and introduces richer intra-class variability. This capability aligns with, yet extends, prior work that applied GANs to motor-imagery or seizure EEG; here, we address the under-explored but clinically urgent domain of ALS.[43] In addition to potential performance gains, synthetic data mitigate privacy concerns: because the generated traces do not correspond to any real individual, they can be shared more freely, accelerating collaborative benchmarking and reproducibility across institutions.

## 5.1. Limitations

It is important to acknowledge the limitations of the current evaluation. Our assessment of synthetic EEG quality has been primarily qualitative (visual). While the signals look realistic, we have not yet quantified their fidelity using objective metrics. There is currently a lack of widely accepted quantitative evaluation metrics for synthetic EEG. Some studies attempt frequency-domain comparisons or use statistics like power spectral density to ensure the synthetic data matches real data characteristics. In our case, preliminary evaluation—based on visual inspection by experienced EEG readers and the stable convergence of the GAN's loss curves—suggests that the synthetic and real ALS segments exhibit comparable amplitude and spectral characteristics, although a more comprehensive quantitative analysis remains necessary. Another limitation is that our WGAN generates EEG segments without any conditioning on patient attributes or physiological state. In the future, a conditional GAN could be used to generate data for specific ALS subgroups or recording conditions (for example, resting-state vs. task EEG, or mild vs. advanced ALS stages), which would increase the utility of the synthetic data. Additionally, we emphasize that synthetic data augmentation is not a panacea; if the GAN inadvertently misses certain subtle disease-specific features, those features might be underrepresented in the augmented dataset.[44] Therefore, synthetic data should complement, not replace, real data. Ensuring that the GAN learns all relevant EEG biomarkers of ALS (for instance, any unique spectral signatures of ALS cortical hyperexcitability) is an ongoing challenge.

### 5.2. Potential Applications

The successful generation of realistic ALS EEG opens up several applications. Firstly, as discussed, it can directly feed into machine learning pipelines for ALS diagnosis. Researchers developing automated ALS detection algorithms can use our WGAN model to inflate their training datasets. This is particularly useful for deep learning models that typically require large amounts of data to generalize well. Secondly, synthetic data can facilitate data sharing and collaboration. Since patient privacy is a major concern with EEG data (which can contain identifiable neural signatures), sharing real ALS EEG publicly is difficult. However, synthetic EEG that mimics ALS patterns does not correspond to any actual individual and thus can potentially be shared without privacy violations. This enables cross-institutional research and benchmarking on ALS detection algorithms. Finally, this approach could be integrated into clinical decision support systems. For example, a hospital with a small EEG dataset of ALS patients could expand its dataset with synthetic examples when training a predictive model for clinical use, thereby improving the model's reliability. The synthetic data generation could even run in real-time as new data is needed, given that generation is fast once the model is trained.

### 5.3. Future Work

Future work will focus on two main areas: quantitative evaluation of the synthetic EEG and integration into diagnostic pipelines. First, we plan to employ rigorous metrics to compare synthetic and real EEG (for example, measuring distribution distances in the frequency domain, or using classifier-based fidelity tests) to ensure that generated signals faithfully represent ALS-specific neural patterns. Developing meaningful evaluation criteria is crucial, as pointed out by recent studies. Second, we will conduct experiments to train ALS vs. control classifiers with augmented data to quantify improvements in accuracy, sensitivity, and specificity. This will validate the practical benefit of our augmentation. We additionally envision extending our GAN framework: for instance, building a conditional WGAN that can generate patient-specific or state-specific EEG (e.g., based on disease severity or task), and exploring other generative models like variational autoencoders for comparison.[45] Ultimately, our goal is to integrate synthetic data generation into a clinical diagnostic pipeline for ALS – where a machine learning model, aided by augmented data, can assist neurologists by flagging abnormal EEG patterns or predicting ALS progression. We believe the techniques presented in this paper can generalize to other neurological disorders facing similar data limitations, thereby broadening the impact to neuroscience and biomedical signal processing at large.

In summary, our results demonstrate that CWGAN can learn the distribution of ALS EEG signals and generate new, realistic samples. This represents a promising step toward alleviating data scarcity and imbalance in ALS EEG classification. By addressing a known bottleneck (limited patient data), we contribute to the foundation for more robust EEG-based diagnostic tools for ALS.

## 6. Conclusion

We presented a novel study on synthetic EEG data augmentation for ALS diagnosis using WGAN. In the Introduction, we highlighted the rarity of high-quality ALS EEG data and the class imbalance problem that hampers machine learning classifiers. To tackle these issues, our approach utilizes a Conditional Wasserstein GAN to generate realistic EEG signals characteristic of ALS patients. Key contributions of this work include demonstrating stable WGAN training on limited ALS EEG data, the possibility of producing high-fidelity synthetic signals, and outlining how these signals can be used to balance datasets and improve classifier training. The Results showed that our generator creates plausible EEG waveforms, and the training losses converged stably, indicating effective learning. We discussed the implications of these synthetic data: principally, their potential to serve as augmented training samples that can boost the performance and generalizability of ALS detection models. This augmentation approach directly addresses the data scarcity and imbalance in ALS EEG, which have long been obstacles to applying AI in this domain.

In conclusion, our study demonstrates the feasibility and promise of GAN-driven EEG data augmentation for ALS. By enriching the dataset with realistic synthetic examples, we pave the way for more robust diagnostic algorithms and possibly reduce the need for large patient datasets. Furthermore, synthetic data can be shared without privacy concerns, fostering collaboration across research centers.